\documentclass[journal]{IEEEtran}
\usepackage[pdftex]{graphicx}
\usepackage{amsmath}
\usepackage{amssymb}
\usepackage{textcomp,booktabs}
\usepackage{multirow}
\usepackage[usenames]{color}

\begin{document}
%

\title{Exploring Rich and Efficient Spatial Temporal Interactions for Real Time Video Salient Object Detection}


\author{Chenglizhao Chen$^{1}$
~~~~Guotao Wang$^{1}$ ~~~~Chong Peng$^{1*}$\thanks{Corresponding
author: Chong Peng (pchong1991@163.com), Chenglizhao Chen and Guotao Wang
contributed equally to this work.}~~~Yuming Fang$^{2}$~~~Dingwen Zhang$^{3}$~~~Hong Qin$^{4}$\\ $^1$Qingdao University ~~$^2$ Jiangxi University of Finance and Economics\\
$^3$Xidian University$~~^4$Stony Brook University \\
Code\&Data: {https://github.com/guotaowang/STVS}\\
}

\markboth{IEEE Transactions on Image Processing, VOL.XX, NO.XX, XXX.XXXX}%
{Shell \MakeLowercase{\textit{et al.}}: Bare Demo of IEEEtran.cls for Journals}

\maketitle

\IEEEtitleabstractindextext{
\begin{abstract}
The current main stream methods formulate their video saliency mainly from two independent venues, i.e., the spatial and temporal branches.
As a complementary component, the main task for the temporal branch is to intermittently focus the spatial branch on those regions with salient movements.
In this way, even though the overall video saliency quality is heavily dependent on its spatial branch, however, the performance of the temporal branch still matter.
Thus, the key factor to improve the overall video saliency is how to further boost the performance of these branches efficiently.
In this paper, we propose a novel spatiotemporal network to achieve such improvement in a full interactive fashion.
We integrate a lightweight temporal model into the spatial branch to coarsely locate those spatially salient regions which are correlated with trustworthy salient movements.
Meanwhile, the spatial branch itself is able to recurrently refine the temporal model in a multi-scale manner.
In this way, both the spatial and temporal branches are able to interact with each other, achieving the mutual performance improvement.
Our method is easy to implement yet effective, achieving high quality video saliency detection in real-time speed with 50 FPS.

\end{abstract}
}
\maketitle
\IEEEdisplaynontitleabstractindextext
\IEEEpeerreviewmaketitle

\begin{figure*}[ht]
\centering
\includegraphics[width=1\linewidth]{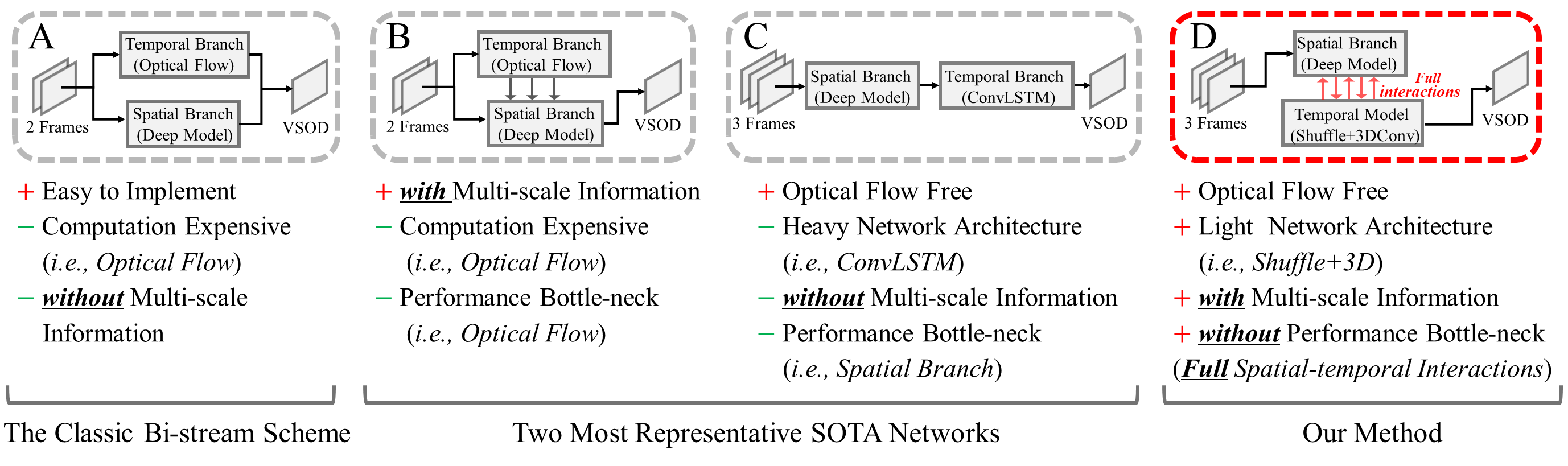}
\caption{The demonstration of the structure differences between our method and the SOTA methods; VSOD: Video Salient Object Detection; sub figure A demonstrates the early classic bi-stream structure; sub figure B and C are the two current most representative structures (e.g., B:~\cite{li2019motion}, C:~\cite{fan2019shifting}), which bias its video saliency to their spatial branch; sub figure D is our novel structure, which make full use of the multi-scale spatiotemporal information for an extremely fast video saliency detection; Also, we have summarized the advantages ($\textcolor[rgb]{1.000, 0.000, 0.000}{\textbf{+}}$) and disadvantages ($\textcolor[rgb]{0.000, 1.000, 0.000}{\textbf{-}}$) below each sub figures.}
\label{fig:Motivation}
\end{figure*}

\section{Introduction and Motivation}

The problem of video salient object detection aims to extract the most visually distinctive objects in video data.
Previous works~\cite{wang2015consistent,liu2016saliency,Chen2019TIP,CC2019TMM2} generate their video saliency maps mainly by fusing the saliency cues which are respectively revealed from its spatial and temporal branches.
In such case, the main task for its spatial branch is to estimate the color saliency cue in single frame, yet its temporal branch aims the motion saliency cue between consecutive multiple frames.

After entering the deep learning era, we can use the off-the-shelf image saliency deep models~\cite{CC2019TMM1,OurTIP15,CC2020TIP} to serve as the spatial branch, and thus we have no intention to give a detailed introduction on this beyond scope issue here.
On the other hand, most of the previous works have adopted the Optical Flow to sense the motion cues, and these motion cues will latterly be feeded into the temporal branch to coarsely locate those regions with salient movements~\cite{ChenPR16,OurSPL18}.
However, the heavy computational optical flow is the major performance bottle-neck, see the pictorial demonstration in the left part of Fig.~\ref{fig:Motivation}.

To further improve, the recent works have focused their video saliency on the spatial branch, while the temporal branch becomes a subordinate to the spatial branch, aiming to intermittently shrink the spatial problem domain~\cite{song2018pyramid,fan2019shifting,li2019motion}.
However, for video data, the color information is frequently more stable than the motion information, e.g., the motion information may be absent completely if the salient
object stay static without any movement for a long period~\cite{ChenPR16}.
Thus, instead of using the “full interaction” strategy,
the SOTA deep learning based methods either follow the
“single-direction interaction” which bias their spatiotemporal trade-off to the spatial branch (e.g., MGA~\cite{li2019motion}), or choose
the “single-scale interaction” to enable the fast end-to-end
network training/testing (e.g., SSAV~\cite{fan2019shifting}).
With regard to further clarify these issue, here, we have demonstrated 2 most representative SOTA architectures in the middle column of Fig.~\ref{fig:Motivation}.

The first architecture (sub figure B in Fig.~\ref{fig:Motivation}) continue uses the optical flow as the input of its temporal branch, while it treats its temporal branch provided motion saliency cues as the side-layer attentions to facilitate the spatial deep feature computation in the multi-scale manner~\cite{li2019motion}.
Although such multi-scale spatiotemporal fusion can indeed improve the overall performance, its performance is still limited by its optical flow usage.
Though the FlowNet can conduct the optical flow computation almost real-time, its computational cost is still the major bottle-neck for the extremely fast video saliency detection.
Meanwhile, the the optical flow provided motion information is occasionally inaccurate, which further degenerate the overall robustness.

The second architecture has abandoned the optical flow usage, adopting the end-to-end ConvLSTM network to sense the temporal saliency cues~\cite{song2018pyramid,fan2019shifting}.
As a subsequent component to the spatial branch, the ConvLSTM network takes the output of its precedent spatial branch as input, seeking the consistent spatial saliency over the time scale as the spatiotemporal video saliency, see sub figure C in Fig.~\ref{fig:Motivation}.
Despite its merit of fast computation, such single stream structure has the following limitations:
\underline{1)} The ConvLSTM network is heavily dependent on its precedent spatial branch, which easily lead to performance bottle-neck when the given video data is dominated by the motion cues;
\underline{2)} Due to the heavy network architecture of the ConvLSTM, the overall video saliency computation speed is limited to about 10 FPS;
\underline{3)} Its temporal branch (ConvLSTM) can not make full use of the multi-scale spatial deep features of its precedent spatial branch, while these two branches should interact with each other to boost their performance mutually~\cite{fang2014video,le2018video}.

Thus, in this paper, we propose a novel spatiotemporal network for video saliency detection, and its key component is the newly designed temporal model to sense the temporal information in an extremely fast speed, see sub figure D in Fig.~\ref{fig:Motivation}.
In sharp contrast to all previous works, our method attempts to be “full interaction” between spatial and temporal branches. The reason is that our temporal module is lightweight designed yet with strong temporal ability, which can be directly inserted into each UNet decoder layer, receiving “multi-scale spatial information” to boost its robustness, whereas the conventional temporal branches can not receive such
“multi-scale spatial information” (e.g., ConvLSTM).
Our spatiotemporal network receive 3 consecutive video frames each time, in which its temporal model mainly consists of simple operations, i.e., sequential 3D convolutions with temporal shuffle operations.
Benefited by such a lightweight designation, it is feasible to integrate the temporal model into each spatial feature layer.
In this way, we can make full use of the multi-scale spatial deep features while sensing the motion saliency cues over temporal scale.
Meanwhile, as an additional convolutional part for each spatial feature layer, the temporal model is able to facilitate the spatial deep feature computation in a recurrent manner.

In summary, compared to the current SOTA methods, our spatiotemporal network has three prominent advantages:
\begin{itemize}
  \item Instead of using the time-consuming optical flow for temporal information, we have devised a novel temporal module, which is very fast and compatible with the current main-stream encoder-decoder structures, achieving real-time
detection with 50 FPS;
      \vspace{0.2cm}
  \item We have inserted the novel temporal module into the decoder layers of vanilla UNet, which can take full advantage of the multi-scale spatial information to alleviate the temporal module induced object boundary blur;
      \vspace{0.2cm}
  \item Moreover, because our temporal branch is more accurate than previous works, we can use the feature propagated from the temporal branch to further improve the performance of our spatial branch, achieving the full spatiotemporal interactions.
\end{itemize}

\begin{figure*}[ht]
\centering
\includegraphics[width=1\linewidth]{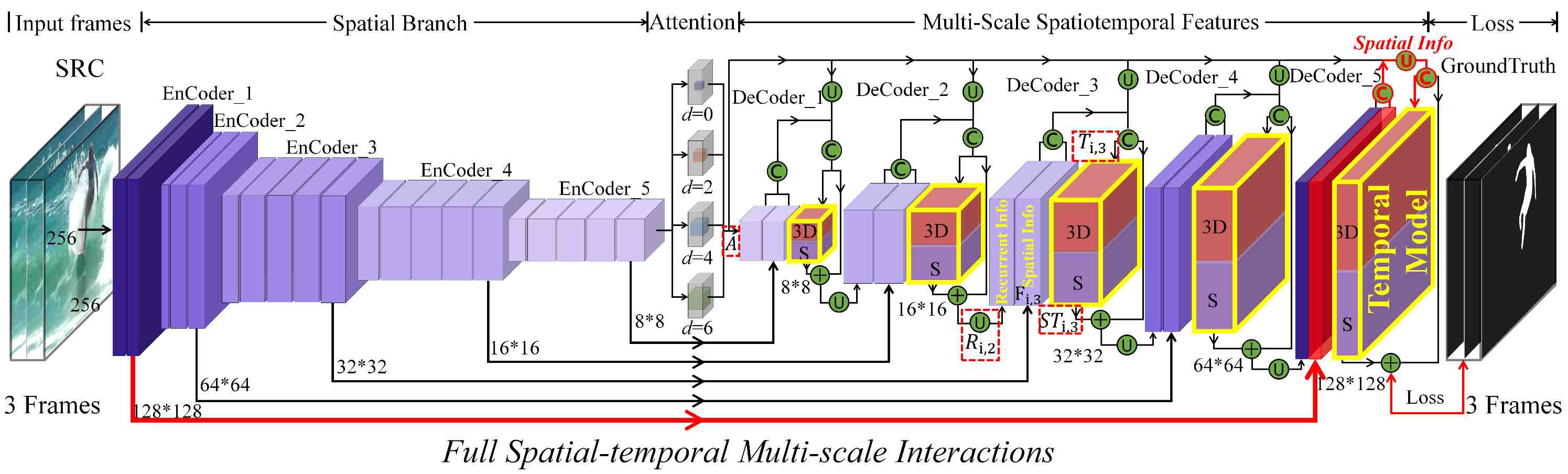}
\caption{The network architecture overview of our spatiotemporal network; we choose the vanilla UNet~\cite{ronneberger2015u} encoder-decoder structure as our baseline network (Sec.~\ref{sec:Flow}); the major highlight of our network is that we have assigned one temporal model, which consists of sequential 3D Convolutions (3D, Sec.~\ref{sec:3D}) with Temporal Shuffle Scheme (S, Sec.~\ref{sec:S}), for each decoder layer to achieve an extremely fast multi-scale spatiotemporal feature computation; \{U,C,$+$\} respectively denote feature up-sample, feature concatenation, feature element-wise summation.}
\label{fig:Net}
\end{figure*}

\section{Related Work}

\subsection{Handcraft Feature Based Methods}
Conventional methods estimate the temporal information mainly using the contrast computation over the hand-crafted features.
Liu~\emph{et al}.~\cite{liu2014superpixel} computed superpixel-wise spatiotemporal contrast histogram to sense the temporal information. Similarly, Liu~\emph{et al}.~\cite{liu2016saliency} proposed the intra-frame similarity matrices to perform the bi-directional temporal propagation as the temporal information.
Wang~\emph{et al}.~\cite{wang2015saliency} acquired the spatiotemporal feature by computing robust geodesic measurement for locating spatial edges and motion boundaries.
~\cite{wang2015consistent} adopted the gradient flow field to obtain intra-frame boundary information as well as inter-frame motion information.
Further, Chen~\emph{et al}.~\cite{chen2018bilevel,chen2017video} adopted the long-term information to enhance the spatiotemporal saliency consistency.

\subsection{Deep Learning Based Methods}
Benefited by the development of deep learning techniques, the current SOTA methods have widely adopted the bi-stream network structure, in which one of its streams aims the color saliency over the spatial information, and another stream extracts the motion saliency over the temporal scale.
Le \emph{et al}.~\cite{le2018video} has adopted such bi-stream structure, in which one of its streams computes the superpixel-wise spatial saliency cues, and another one aims the temporal saliency computation by applying the 3D convolution directly over multiple video frames.
However, its direct usage of 3D convolution has totally overlooked the spatial information, obscuring its detection boundary severely.
Wang \emph{et al}.~\cite{wang2017video} proposed to use 2D convolution network to sense the differences between adjacent two frames as the temporal information, in which its behind rationale is quiet similar to the deep learning based optical flow computation.
Song \emph{et al}.~\cite{song2018pyramid} adopted several dilated ConvLSTMs to extract multi-scale spatiotemporal information and feed it into bi-directional ConvLSTM to obtain the spatiotemporal information.
Wang \emph{et al}.~\cite{wang2019learning}, Fan et al ~\cite{fan2019shifting} further adopted the human visual fixation to enable its video saliency shifting between different objects.
Li \emph{et al}.~\cite{li2018flow} has adopted the FlowNet based optical flow to sense the temporal information, and then these temporal information will be used to enhance the saliency consistency over temporal scale by using the ConvLSTM network.
Most recently, Li \emph{et al}.~\cite{li2019motion} have developed a multi-task network.
As usually, it has adopted the optical flow to sense the temporal information, in which its major highlight is that it utilizes its temporal branch to accomplish its spatial branch, achieving a significant performance improvement.
In ~\cite{li2019motion}, though its temporal branch is able to affect its spatial branch, it has overlooked the usage of its spatial branch to interact with its temporal branch, in which the color saliency obtained by the spatial branch can indeed affect the temporal branch positively.

\section{Spatiotemporal Network Overview}
\label{sec:Flow}
The classic UNet~\cite{ronneberger2015u} adopts the encoder-decoder network structure, which has a remarkable learning ability to simultaneously extract both high-level semantic information and low-level tiny spatial details.
Thus, we choose it as the baseline network, and its overall network architecture can be found in Fig.~\ref{fig:Net}).

In the case of single image, the high-level semantic information in UNet tends to decrease with the increase of the decoder layers.
However, the problem is that the performance of these increased decoder layers may get degenerated due to the reduced high-level semantic information.
To alleviate it, the widely used scheme is to integrate those deep features in each encoder layer, which is supposed to have abundant of high-level semantic information, into each decoder layer.

In the case of video data, we input 3 video frames into our baseline network each time.
For each encoder layer, we represent its feature block as ${F}_{i,j}, i\in\{1,2,3\}$, where $j$ is the encoder layer index, and $i$ indicates those spatial deep features of the 3 input images.
As shown in Fig.~\ref{fig:Net}, we respectively assign one temporal model (marked by yellow rectangle) for each decoder layer.
The temporal model takes the spatial deep features ${F}_{i,j}$ as input, aiming to reveal an additional high-level semantic information over the temporal scale (i.e., between the 3 input frames).
With the help of these temporal related high-level semantic information, the spatial deep features in the decoder layers can get improved significantly, and thus we name these fused deep features as the spatiotemporal deep features, which is able to sense both the spatial and temporal saliency cues at the same time.

Meanwhile, to make full use of both the multi-scale spatial deep features and the high-level attention based spatial localization information, the temporal model also takes the spatial attention maps as input, and these attention maps ($A$) are computed by applying the dilated convolutions (with dilation factor $d=\{0,2,4,6\}$) over the spatial deep features of the last encoder layer (i.e., ${F}_{i,5}$).
Thus, the computation procedure toward the spatiotemporal deep feature (${ST}_{i,3}$) in the 3rd encoder layer can be detailed as Eq.~\ref{eq:STFeature}.
\begin{equation}
\label{eq:STFeature}
{ST}_{\emph{i},3} = \emph{TM}\Bigg(\emph{Conv}\Big({F}_{\emph{i},3}\otimes \emph{U}({A})\Big)\Bigg),
\end{equation}
where $Conv$ denotes the feature convolutional operation, $\otimes$ is the feature concatenate operation, $TM$ denotes the temporal model, $U$ denotes the up-sample operation.

Moreover, the spatiotemporal deep features ($ST$) will be recurrently integrated into the next decoder layer for robust temporal deep feature computation.
Thus, the complete spatiotemporal deep feature computation procedure should be updated to Eq.~\ref{eq:STFeatureFull}.
\begin{equation}
\label{eq:STFeatureFull}
{ST}_{\emph{i},3} = \emph{TM}\Bigg(\emph{Conv}\Big(\emph{Conv}({F}_{\emph{i},3}\otimes R_{\emph{i},2})\otimes \emph{U}({A})\Big)\Bigg),
\end{equation}
\begin{equation}
\label{eq:R}
{R}_{i,2} = {U\Bigg({\ {ST}_{\emph{i},2}\ }+\ }{{\emph{Conv}\Big({Conv}({F}_{\emph{i},2}\otimes R_{\emph{i},1})\otimes \emph{U}({A})\Big)}}\Bigg),
\end{equation}
where $R_{i,2}$ represents the recurrent features from its previous decoder layer, which can be obtained by Eq.~\ref{eq:R}.
In this way, we have achieved the full interactive status between the spatial branch (i.e., the encoder layers of UNet) and the temporal model.

\begin{figure}[t]
\centering
\includegraphics[width=1\linewidth]{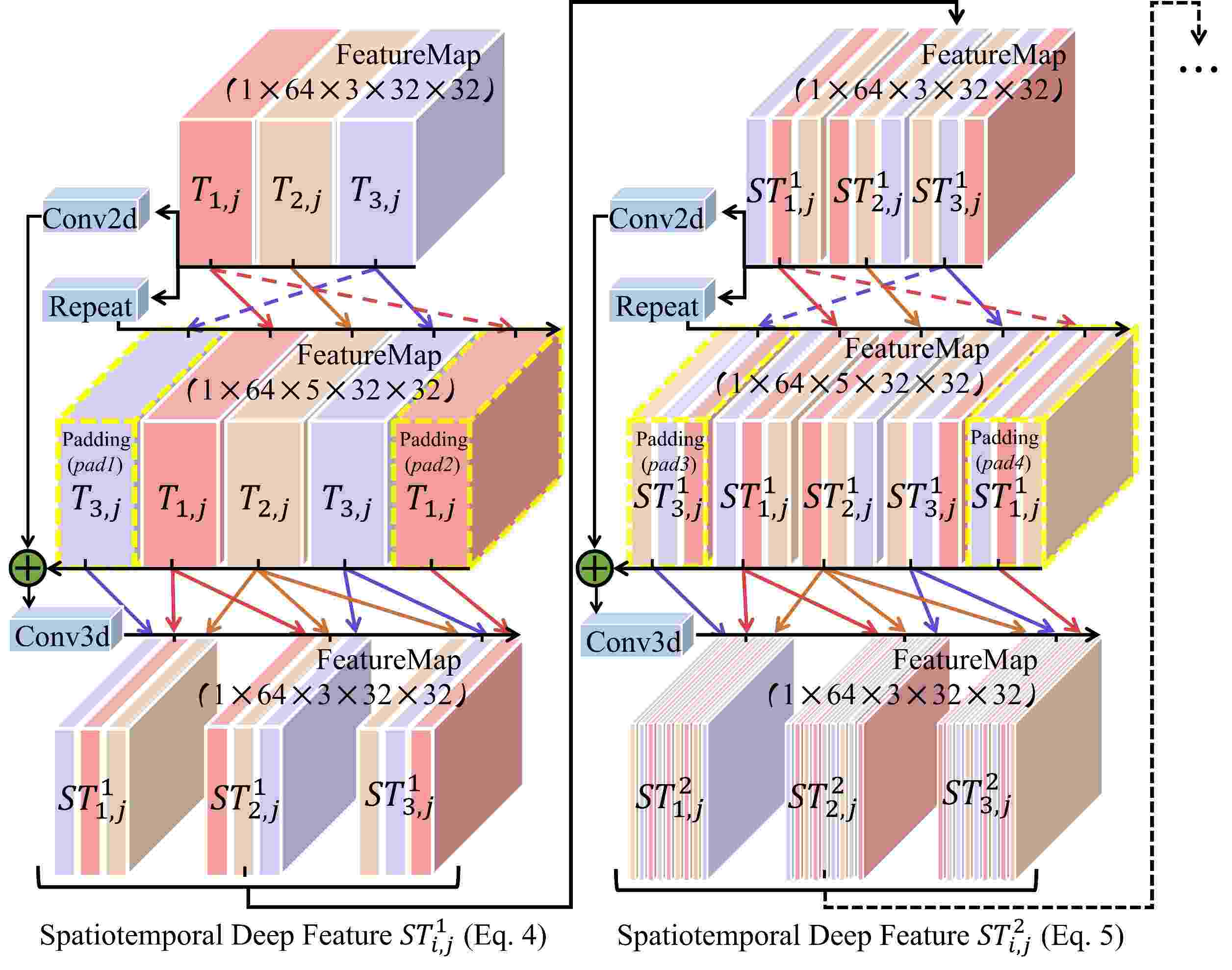}
\vspace{-0.2cm}
\caption{Overall dataflow in temporal model; $T$ and $ST$ respectively denote the input and output of the temporal model; the cyclic padding data is marked by yellow dash line; the repeat operation $+$ sliding window scheme both ensure an extremely fast cyclic padding operation, see Eq.~\ref{eq:repeat}, and more details can be found at the end of Sec.~\ref{sec:3D}.}
\vspace{-0.4 cm}
\label{fig:RNet}
\end{figure}
\section{Our Temporal Model}
\label{sec:TM}
\vspace{-0.1cm}
\subsection{Fast 3D Convolution}
\label{sec:3D}
\vspace{-0.1cm}
Compared to the conventional 2D convolution (e.g., a $3\times3$ flat kernel), which can only sense the spatial information in single video frame, the 3D convolution (e.g., a $3\times3\times3$ cubic kernel) is able to sense the temporal information.
As a basic computational unit, our 3D convolution
itself is exactly the same as the plain 3D convolution, yet
the major difference is how we use it to capture temporal
information. In general, the single plain 3D convolution
frequently with limited temporal sensing ability, however,
our 3D convolution with “fast cyclic padding” scheme kills two birds with one stone:\\
\underline{1)} It avoids the “conventional
padding” induced performance degradation when using sequential 3D convolutions to enhance the temporal sensing
ability; \\
\underline{2)} It cyclicly uses other frames as the padding data,
enhancing the temporal ability naturally without additional
computational costs.

Therefore, for each encoder spatial layer, we use the 3D convolution to sense its spatial common consistency over the temporal scale as the spatiotemporal deep features, e.g., the ${ST}_{\emph{i},3}$ in Eq.~\ref{eq:STFeatureFull}, which is also marked by red dash line in Fig.~\ref{fig:Net}.
As we have mentioned in Eq.~\ref{eq:STFeatureFull}, the input data of our temporal model consists of 3 aspects, including the multi-scale spatial deep features ($F$), the attention maps ($A$), and the recurrent data from the precedent decoder layer ($R$).
For simplicity, we use $T_{i,j}, i\in\{1,2,3\}$ to denote the input deep features (i.e., 3 frames) of the temporal model in the $j$-th decoder layer.
To reveal the temporal information, we use the sliding window scheme to apply the 3D convolution ($Conv3D$) over $T_i$, and thus the temporal model output ($ST_{i,j}$) can be fast computed by Eq.~\ref{eq:3D}.
\begin{equation}
\label{eq:3D}
ST_{i,j}^1=\left\{\begin{array}{lll} Conv3D(pad_1,T_{1,j},T_{2,j})\ \ \ \ &if\ \ i=1\\ Conv3D(T_{1,j},T_{2,j},T_{3,j})\ \ \ \   &if\ \ i=2\\ Conv3D(T_{2,j},T_{3,j},pad_2)\ \ \ \ &if\ \ i=3\end{array}\right.,
\end{equation}
where $pad$ denote the padding data.
Thus far, we have applied the 3D convolution over spatial deep features to sense temporal information, however, we found that only using single 3D convolution is incapable to obtain temporal information accurately.
Thus, in our implementation, we use 3 sequential 3D convolutions in our temporal model, and thus the spatiotemporal deep features of the 2rd 3D convolution can be updated by Eq.~\ref{eq:3D2}.
\begin{equation}
\label{eq:3D2}
ST_{i,j}^2\gets\left\{\begin{array}{lll} Conv3D(pad_3,ST_{1,j}^1,ST_{2,j}^1)\ &if\ \ i=1\\ Conv3D(ST_{1,j}^1,ST_{2,j}^1,ST_{3,j}^1)\ &if\ \ i=2\\ Conv3D(ST_{2,j}^1,ST_{3,j}^1,pad_4)\  &if\ \ i=3\end{array}\right..
\end{equation}

As shown in Eq.~\ref{eq:3D} and Eq.~\ref{eq:3D2}, there exists 2 major problems toward the sequential usage of multiple 3D convolutions:\\
\underline{1)} Due to the miss-aligned spatial information between different video frames, the direct usage of multiple 3D convolution easily lead to loss the tiny spatial details, blurring the object boundaries in its detection results;\\
\underline{2)} The intuitive padding scheme (e.g., zero padding) may lead the computed spatiotemporal deep features problematical after using multiple sequential 3D convolutions.

To solve the problem \underline{1)}, we simply add each $ST$ with the deep features computed by 2D convolution, e.g., $ST^1_{1,j}\gets \{ST^1_{1,j}+Conv(T_{1,j})\}$, $ST^2_{1,j}\gets \{ST^2_{1,j}+Conv(ST^1_{1,j})\}$.
Meanwhile, we use the cyclic padding scheme to handle the problem \underline{2)}, see Eq.~\ref{eq:pad}.
\begin{equation}
\label{eq:pad}
\{pad_1,pad_2,pad_3,pad_4\}\gets \{T_{3,j},T_{1,j},ST^1_{1,j},ST^1_{3,j}\}.
\end{equation}
However, such cyclic padding scheme is time consuming if we reorganize the input data.
Therefore, we directly use the repeat operation on GPU 3 times to expand the original $\{T_1,T_2,T_3\}$ into $\{T_1,T_2,T_3,T_1,T_2,T_3,T_1,T_2,T_3\}$, and thus the cyclic padding can be fulfilled by using a sliding window over these expanded features, and such implementation is 5 times faster than the conventional feature reorganization, see the example in Eq.~\ref{eq:repeat}.
\begin{equation}
\label{eq:repeat}
\begin{split}
repeat(\{T_1,\underset{ST^1_{1,j}=Conv3D(...)\ \  }{T_2,T_3\},3)\rightarrow\{T_1,T_2,\underbrace{T_3,T_1,T_2},T_3,T_1,...}\}.
\end{split}
\vspace{-0.2cm}
\end{equation}
Also, we have shown the complete data flow of our temporal model in Fig.~\ref{fig:RNet}

\subsection{Fast Temporal Shuffle}
\label{sec:S}
The temporal information sensed by the 3D convolutions is mainly from the 3rd dimension of the adopted 3D kernels (e.g., ``Spatial'':$\{3\times3\}\times$``Temporal'':$\{3\}$), which may lead the final spatiotemporal deep features bias to the spatial domain, degenerating the performance of the temporal model.
To solve it, we propose a fast temporal shuffle scheme to enhance the temporal information sensing ability of the temporal model.

Inspired by the ShuffleNet~\cite{zhang2018shufflenet} which enhances its spatial feature diversity by randomly scrambling its feature orders, here, we enhance the temporal part in our temporal model by swapping the deep features between consecutive video frames.
For an example, as shown in Eq.~\ref{eq:shuffle}, we swap the deep feature $f_a$ in frame 1 (i.e., $ST_1$) with the deep feature $f_b$ in frame 2 (i.e., $ST_2$), and we swap $f_c$ with $f_d$ as well.
\begin{equation}
\label{eq:shuffle}
ST_1\{...,\underset{swap}{\underbrace{f_a,...\}\Big\vert ST_2\{...,f_b}},...,\underset{swap}{\underbrace{f_c,...\}\Big\vert ST_3\{...,f_d}},...\}.
\end{equation}

\begin{figure}[t]
\centering
\includegraphics[width=1\linewidth]{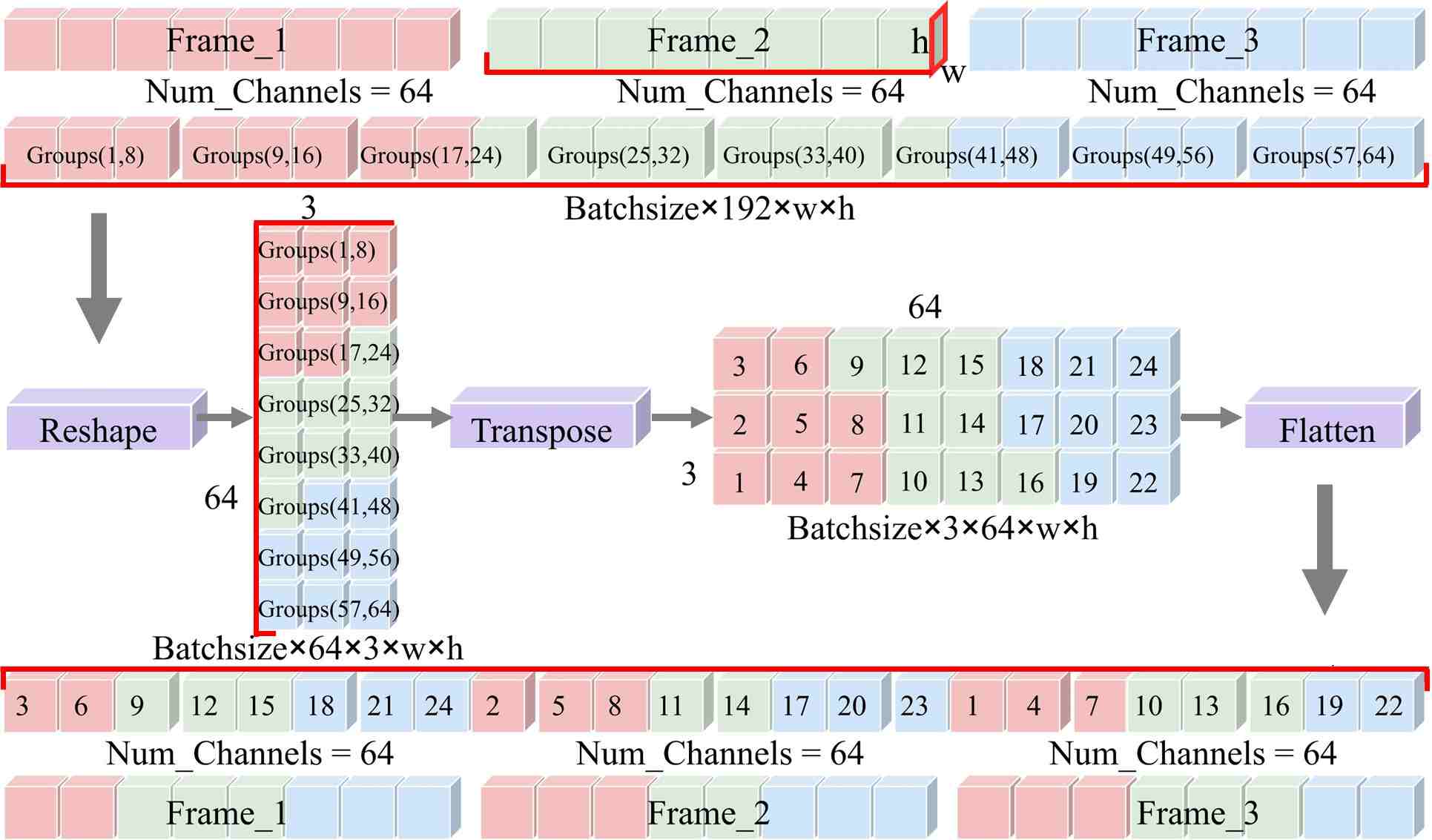}
\caption{The detailed illustration of our fast temporal shuffle scheme. The Top $Frame\_i, i\in\{1,2,3\}$ and Bottom $Frame\_i, i\in\{1,2,3\}$ respectively represent the input and output features of 3 consecutive video frames, of which the input deep features (64 channels) pass through each other via our fast temporal shuffle (see details in Sec.~\ref{sec:S}).}
\label{fig:SNet}
\end{figure}

In fact, the above temporal shuffle can be fully implemented in GPU, which can be extremely simple and fast, see the pictorial demonstration in Fig.~\ref{fig:SNet}.
We first sequentially divide the original $192\times1$ deep features into 64 groups, and each group includes 3 deep features.
Next, we reshape the original $192\times1$ deep features into $64\times3$ according to their group orders, and then we transpose it into $3\times64$, and finally we flatten it back to $192\times1$ deep features.
In this way, we have automatically inserted the temporally neighbored spatial features into the current frame.

In our implementation, we repeat the above temporal shuffle 2 times, i.e., one for the output ($ST^1$) of the first 3D convolution , and another for the output ($ST^2$) of the second 3D convolution.
Thus the complete dataflow in temporal model can be represented by Eq.~\ref{eq:OverallDataflow}.
\begin{equation}
\label{eq:OverallDataflow}
ST\gets Conv3D\Bigg(S\bigg(Conv3D\Big(S\big(Conv3D(T)\big)\Big)\bigg)\Bigg),
\end{equation}
where $T$ and $ST$ respectively denote the input and output of the temporal model, $S$ represents the temporal shuffle operation.

\begin{table*}[htbp]
\centering
  \caption{Quantitative comparisons between our method and the SOTA video saliency detection methods using F-max, S-measure, MAE metrics, the top three results are respectively highlighted in \textcolor[rgb]{1.00,0.00,0.00}{red}, \textcolor[rgb]{0.00,1.00,0.00}{green}, \textcolor[rgb]{0.00,0.00,1.00}{blue}, "---"indicate the model trained on this datasets, "*" indicates the model was deep learning based video saliency detection methods, and "**" indicates the results was deep learning based image saliency detection methods.}
  \resizebox{\textwidth}{!}{
  \LARGE
    \begin{tabular}{l|c|c|c|c|c|c|c|c|c|c|c|c|c|c|c|c|c|c|c}
    \toprule[1.5pt]
    DataSets & -     & \multicolumn{3}{c|}{DAVIS-T~\cite{perazzi2016benchmark}} & \multicolumn{3}{c|}{SegTrack-V2~\cite{li2013video}} & \multicolumn{3}{c|}{ViSal~\cite{wang2015consistent}} & \multicolumn{3}{c|}{FBMS-T~\cite{ochs2013segmentation}} & \multicolumn{3}{c|}{VOS-T~\cite{li2017benchmark}} & \multicolumn{3}{c}{DAVSOD-T~\cite{fan2019shifting}} \\
    \midrule
    Metric & Year  & \multicolumn{1}{c}{F-max} & \multicolumn{1}{c}{S-measure} & MAE   & \multicolumn{1}{c}{F-max} & \multicolumn{1}{c}{S-measure} & MAE   & \multicolumn{1}{c}{F-max} & \multicolumn{1}{c}{S-measure} & MAE   & \multicolumn{1}{c}{F-max} & \multicolumn{1}{c}{S-measure} & MAE   & \multicolumn{1}{c}{F-max} & \multicolumn{1}{c}{S-measure} & MAE   & \multicolumn{1}{c}{F-max} & \multicolumn{1}{c}{S-measure} & MAE \\
    \midrule
    \midrule
    \textbf{OUR} & -     & \multicolumn{1}{c}{\textcolor[rgb]{0.000, 0.000, 1.000}{\textbf{0.865}}} & \multicolumn{1}{c}{0.892} & \textcolor[rgb]{1.000, 0.000, 0.000}{\textbf{0.023}} & \multicolumn{1}{c}{\textcolor[rgb]{1.000, 0.000, 0.000}{\textbf{0.860}}} & \multicolumn{1}{c}{\textcolor[rgb]{1.000, 0.000, 0.000}{\textbf{0.891}}} & \textcolor[rgb]{1.000, 0.000, 0.000}{\textbf{0.017}} & \multicolumn{1}{c}{\textcolor[rgb]{0.000, 0.580, 0.231}{\textbf{0.952}}} & \multicolumn{1}{c}{\textcolor[rgb]{0.000, 0.580, 0.231}{\textbf{0.952}}} & \textcolor[rgb]{1.000, 0.000, 0.000}{\textbf{0.013}} & \multicolumn{1}{c}{\textcolor[rgb]{0.000, 0.000, 1.000}{\textbf{0.856}}} & \multicolumn{1}{c}{0.872} & \textcolor[rgb]{0.000, 0.000, 1.000}{\textbf{0.038}} & \multicolumn{1}{c}{\textcolor[rgb]{1.000, 0.000, 0.000}{\textbf{0.791}}} & \multicolumn{1}{c}{\textcolor[rgb]{1.000, 0.000, 0.000}{\textbf{0.850}}} & \textcolor[rgb]{1.000, 0.000, 0.000}{\textbf{0.058}} & \multicolumn{1}{c}{\textcolor[rgb]{0.000, 0.580, 0.231}{\textbf{0.651}}} & \multicolumn{1}{c}{\textcolor[rgb]{0.000, 0.580, 0.231}{\textbf{0.746}}} & \textcolor[rgb]{0.000, 0.580, 0.231}{\textbf{0.086}} \\
    \midrule
    MGA~\cite{li2019motion}* & 2019  & \multicolumn{1}{c}{\textcolor[rgb]{1.000, 0.000, 0.000}{\textbf{0.892}}} & \multicolumn{1}{c}{\textcolor[rgb]{1.000, 0.000, 0.000}{\textbf{0.910}}} & \textcolor[rgb]{1.000, 0.000, 0.000}{\textbf{0.023}} & \multicolumn{1}{c}{\textcolor[rgb]{0.000, 0.580, 0.000}{\textbf{0.821}}} & \multicolumn{1}{c}{\textcolor[rgb]{0.000, 0.580, 0.000}{\textbf{0.865}}} & 0.030  & \multicolumn{1}{c}{0.933} & \multicolumn{1}{c}{0.936} & 0.017  & \multicolumn{1}{c}{\textcolor[rgb]{1.000, 0.000, 0.000}{\textbf{0.899}}} & \multicolumn{1}{c}{\textcolor[rgb]{1.000, 0.000, 0.000}{\textbf{0.904}}} & \textcolor[rgb]{1.000, 0.000, 0.000}{\textbf{0.028}} & \multicolumn{1}{c}{0.735} & \multicolumn{1}{c}{0.792} & 0.075  & \multicolumn{1}{c}{\textcolor[rgb]{0.000, 0.000, 1.000}{\textbf{0.640}}} & \multicolumn{1}{c}{\textcolor[rgb]{0.000, 0.000, 1.000}{\textbf{0.738}}} & \textcolor[rgb]{1.000, 0.000, 0.000}{\textbf{0.084}} \\
    AGS~\cite{wang2019learning}* & 2019  & \multicolumn{1}{c}{\textcolor[rgb]{0.000, 0.580, 0.231}{\textbf{0.873}}} & \multicolumn{1}{c}{\textcolor[rgb]{0.000, 0.580, 0.231}{\textbf{0.898}}} & \textcolor[rgb]{0.000, 0.580, 0.231}{\textbf{0.026}} & \multicolumn{1}{c}{\textcolor[rgb]{0.000, 0.000, 1.000}{\textbf{0.816}}} & \multicolumn{1}{c}{0.858} & \textcolor[rgb]{0.000, 0.000, 1.000}{\textbf{0.022}} & \multicolumn{1}{c}{\textcolor[rgb]{1.000, 0.000, 0.000}{\textbf{0.960}}} & \multicolumn{1}{c}{\textcolor[rgb]{1.000, 0.000, 0.000}{\textbf{0.960}}} & \textcolor[rgb]{0.000, 0.000, 1.000}{\textbf{0.014}} & \multicolumn{1}{c}{0.840} & \multicolumn{1}{c}{0.874} & 0.048  & \multicolumn{1}{c}{\textcolor[rgb]{0.000, 0.580, 0.231}{\textbf{0.774}}} & \multicolumn{1}{c}{\textcolor[rgb]{0.000, 0.580, 0.231}{\textbf{0.840}}} & \textcolor[rgb]{0.000, 0.580, 0.231}{\textbf{0.066}} & \multicolumn{1}{c}{\textcolor[rgb]{1.000, 0.000, 0.000}{\textbf{0.661}}} & \multicolumn{1}{c}{\textcolor[rgb]{1.000, 0.000, 0.000}{\textbf{0.759}}} & \textcolor[rgb]{0.000, 0.000, 1.000}{\textbf{0.090}} \\
    SSAV~\cite{fan2019shifting}* & 2019  & \multicolumn{1}{c}{0.861} & \multicolumn{1}{c}{\textcolor[rgb]{0.000, 0.000, 1.000}{\textbf{0.893}}} & \textcolor[rgb]{0.000, 0.000, 1.000}{\textbf{0.028}} & \multicolumn{1}{c}{0.801} & \multicolumn{1}{c}{0.851} & 0.023 & \multicolumn{1}{c}{0.939} & \multicolumn{1}{c}{0.943} & 0.020 & \multicolumn{1}{c}{\textcolor[rgb]{0.000, 0.580, 0.231}{\textbf{0.865}}} & \multicolumn{1}{c}{\textcolor[rgb]{0.000, 0.580, 0.000}{\textbf{0.879}}} & 0.040 & \multicolumn{1}{c}{\textcolor[rgb]{0.000, 0.000, 1.000}{\textbf{0.742}}} & \multicolumn{1}{c}{\textcolor[rgb]{0.000, 0.000, 1.000}{\textbf{0.819}}} & 0.073 & \multicolumn{1}{c}{0.603} & \multicolumn{1}{c}{0.724} & 0.092 \\
    \midrule
    CPD~\cite{wu2019cascaded}** & 2019  & \multicolumn{1}{c}{0.778} & \multicolumn{1}{c}{0.859} & 0.032  & \multicolumn{1}{c}{0.778} & \multicolumn{1}{c}{0.841} & 0.023  & \multicolumn{1}{c}{0.941} & \multicolumn{1}{c}{0.942} & 0.016  & \multicolumn{1}{c}{0.810} & \multicolumn{1}{c}{0.846} & 0.048  & \multicolumn{1}{c}{0.735} & \multicolumn{1}{c}{0.818} & \textcolor[rgb]{0.000, 0.000, 1.000}{\textbf{0.068}} & \multicolumn{1}{c}{0.608} & \multicolumn{1}{c}{0.724} & 0.092  \\
    PoolNet~\cite{liu2019simple}** & 2019  & \multicolumn{1}{c}{0.827} & \multicolumn{1}{c}{0.860} & 0.044  & \multicolumn{1}{c}{0.782} & \multicolumn{1}{c}{0.843} & \textcolor[rgb]{0.000, 0.580, 0.231}{\textbf{0.020}} & \multicolumn{1}{c}{\textcolor[rgb]{0.000, 0.000, 1.000}{\textbf{0.945}}} & \multicolumn{1}{c}{0.945} & \textcolor[rgb]{0.000, 0.000, 1.000}{\textbf{0.015}} & \multicolumn{1}{c}{\textcolor[rgb]{0.000, 0.000, 1.000}{\textbf{0.856}}} & \multicolumn{1}{c}{\textcolor[rgb]{0.000, 0.000, 1.000}{\textbf{0.878}}} & \textcolor[rgb]{0.000, 0.580, 0.231}{\textbf{0.037}} & \multicolumn{1}{c}{0.719} & \multicolumn{1}{c}{0.796} & 0.076  & \multicolumn{1}{c}{0.612} & \multicolumn{1}{c}{0.731} & 0.088  \\
    EGNet~\cite{zhao2019egnet}** & 2019  & \multicolumn{1}{c}{0.767} & \multicolumn{1}{c}{0.828} & 0.057  & \multicolumn{1}{c}{0.774} & \multicolumn{1}{c}{0.848} & 0.024  & \multicolumn{1}{c}{0.941} & \multicolumn{1}{c}{\textcolor[rgb]{0.000, 0.000, 1.000}{\textbf{0.946}}} & \textcolor[rgb]{0.000, 0.000, 1.000}{\textbf{0.015}} & \multicolumn{1}{c}{0.848} & \multicolumn{1}{c}{\textcolor[rgb]{0.000, 0.000, 1.000}{\textbf{0.878}}} & 0.044  & \multicolumn{1}{c}{0.698} & \multicolumn{1}{c}{0.793} & 0.082  & \multicolumn{1}{c}{0.604} & \multicolumn{1}{c}{0.719} & 0.101  \\
    \midrule
    PDBM~\cite{song2018pyramid}* & 2018  & \multicolumn{1}{c}{0.855} & \multicolumn{1}{c}{0.882} & \textcolor[rgb]{0.000, 0.000, 1.000}{\textbf{0.028}} & \multicolumn{1}{c}{0.800} & \multicolumn{1}{c}{\textcolor[rgb]{0.000, 0.000, 1.000}{\textbf{0.864}}} & 0.024 & \multicolumn{1}{c}{0.888} & \multicolumn{1}{c}{0.907} & 0.032 & \multicolumn{1}{c}{0.821} & \multicolumn{1}{c}{0.851} & 0.064 & \multicolumn{1}{c}{\textcolor[rgb]{0.000, 0.000, 1.000}{\textbf{0.742}}} & \multicolumn{1}{c}{0.818} & 0.078 & \multicolumn{1}{c}{0.572} & \multicolumn{1}{c}{0.698} & 0.116 \\
    MBNM~\cite{li2018unsupervised}* & 2018  & \multicolumn{1}{c}{0.861} & \multicolumn{1}{c}{0.887} & 0.031 & \multicolumn{1}{c}{0.716} & \multicolumn{1}{c}{0.809} & 0.026 & \multicolumn{1}{c}{0.883} & \multicolumn{1}{c}{0.898} & 0.020 & \multicolumn{1}{c}{0.816} & \multicolumn{1}{c}{0.857} & 0.047 & \multicolumn{1}{c}{0.670} & \multicolumn{1}{c}{0.742} & 0.099 & \multicolumn{1}{c}{0.520} & \multicolumn{1}{c}{0.637} & 0.159 \\
    FGRN~\cite{li2018flow}* & 2018  & \multicolumn{1}{c}{0.783} & \multicolumn{1}{c}{0.838} & 0.043 & \multicolumn{1}{c}{---} & \multicolumn{1}{c}{---} & ---    & \multicolumn{1}{c}{0.848} & \multicolumn{1}{c}{0.861} & 0.045 & \multicolumn{1}{c}{0.767} & \multicolumn{1}{c}{0.809} & 0.088 & \multicolumn{1}{c}{0.669} & \multicolumn{1}{c}{0.715} & 0.097 & \multicolumn{1}{c}{0.573} & \multicolumn{1}{c}{0.693} & 0.098 \\
    DLVS~\cite{wang2017video}* & 2018  & \multicolumn{1}{c}{0.708} & \multicolumn{1}{c}{0.794} & 0.061 & \multicolumn{1}{c}{---} & \multicolumn{1}{c}{---} & ---    & \multicolumn{1}{c}{0.852} & \multicolumn{1}{c}{0.881} & 0.048 & \multicolumn{1}{c}{0.759} & \multicolumn{1}{c}{0.794} & 0.091 & \multicolumn{1}{c}{0.675} & \multicolumn{1}{c}{0.760} & 0.099 & \multicolumn{1}{c}{0.521} & \multicolumn{1}{c}{0.657} & 0.129 \\
    SCNN~\cite{tang2018weakly}* & 2018  & \multicolumn{1}{c}{0.714} & \multicolumn{1}{c}{0.783} & 0.064 & \multicolumn{1}{c}{---} & \multicolumn{1}{c}{---} & ---    & \multicolumn{1}{c}{0.831} & \multicolumn{1}{c}{0.847} & 0.071 & \multicolumn{1}{c}{0.762} & \multicolumn{1}{c}{0.794} & 0.095 & \multicolumn{1}{c}{0.609} & \multicolumn{1}{c}{0.704} & 0.109 & \multicolumn{1}{c}{0.532} & \multicolumn{1}{c}{0.674} & 0.128 \\
    SCOM~\cite{chen2018scom}* & 2018  & \multicolumn{1}{c}{0.783} & \multicolumn{1}{c}{0.832} & 0.048 & \multicolumn{1}{c}{0.764} & \multicolumn{1}{c}{0.815} & 0.030 & \multicolumn{1}{c}{0.831} & \multicolumn{1}{c}{0.762} & 0.122 & \multicolumn{1}{c}{0.797} & \multicolumn{1}{c}{0.794} & 0.079 & \multicolumn{1}{c}{0.690} & \multicolumn{1}{c}{0.712} & 0.162 & \multicolumn{1}{c}{0.464} & \multicolumn{1}{c}{0.599} & 0.220 \\
    \midrule
    SFLR~\cite{chen2017video} & 2017  & \multicolumn{1}{c}{0.727} & \multicolumn{1}{c}{0.790} & 0.056 & \multicolumn{1}{c}{0.745} & \multicolumn{1}{c}{0.804} & 0.037 & \multicolumn{1}{c}{0.779} & \multicolumn{1}{c}{0.814} & 0.062 & \multicolumn{1}{c}{0.660} & \multicolumn{1}{c}{0.699} & 0.117 & \multicolumn{1}{c}{0.546} & \multicolumn{1}{c}{0.624} & 0.145 & \multicolumn{1}{c}{0.478} & \multicolumn{1}{c}{0.624} & 0.132 \\
    SGSP~\cite{liu2016saliency} & 2017  & \multicolumn{1}{c}{0.655} & \multicolumn{1}{c}{0.692} & 0.138 & \multicolumn{1}{c}{0.673} & \multicolumn{1}{c}{0.681} & 0.124 & \multicolumn{1}{c}{0.677} & \multicolumn{1}{c}{0.706} & 0.165 & \multicolumn{1}{c}{0.630} & \multicolumn{1}{c}{0.661} & 0.172 & \multicolumn{1}{c}{0.426} & \multicolumn{1}{c}{0.557} & 0.236 & \multicolumn{1}{c}{0.426} & \multicolumn{1}{c}{0.577} & 0.207 \\
    STBP~\cite{xi2016salient} & 2017  & \multicolumn{1}{c}{0.544} & \multicolumn{1}{c}{0.677} & 0.096 & \multicolumn{1}{c}{0.640} & \multicolumn{1}{c}{0.735} & 0.061 & \multicolumn{1}{c}{0.622} & \multicolumn{1}{c}{0.629} & 0.163 & \multicolumn{1}{c}{0.595} & \multicolumn{1}{c}{0.627} & 0.152 & \multicolumn{1}{c}{0.526} & \multicolumn{1}{c}{0.576} & 0.163 & \multicolumn{1}{c}{0.410} & \multicolumn{1}{c}{0.568} & 0.160 \\
    MSTM~\cite{tu2016real} & 2016  & \multicolumn{1}{c}{0.429} & \multicolumn{1}{c}{0.583} & 0.165 & \multicolumn{1}{c}{0.526} & \multicolumn{1}{c}{0.643} & 0.114 & \multicolumn{1}{c}{0.673} & \multicolumn{1}{c}{0.749} & 0.095 & \multicolumn{1}{c}{0.500} & \multicolumn{1}{c}{0.613} & 0.177 & \multicolumn{1}{c}{0.567} & \multicolumn{1}{c}{0.657} & 0.144 & \multicolumn{1}{c}{0.344} & \multicolumn{1}{c}{0.532} & 0.211 \\
    GFVM~\cite{wang2015consistent} & 2015  & \multicolumn{1}{c}{0.569} & \multicolumn{1}{c}{0.687} & 0.103 & \multicolumn{1}{c}{0.592} & \multicolumn{1}{c}{0.699} & 0.091 & \multicolumn{1}{c}{0.683} & \multicolumn{1}{c}{0.757} & 0.107 & \multicolumn{1}{c}{0.571} & \multicolumn{1}{c}{0.651} & 0.160 & \multicolumn{1}{c}{0.506} & \multicolumn{1}{c}{0.615} & 0.162 & \multicolumn{1}{c}{0.334} & \multicolumn{1}{c}{0.553} & 0.167 \\
    SAGM~\cite{wang2015saliency} & 2015  & \multicolumn{1}{c}{0.515} & \multicolumn{1}{c}{0.676} & 0.103 & \multicolumn{1}{c}{0.634} & \multicolumn{1}{c}{0.719} & 0.081 & \multicolumn{1}{c}{0.688} & \multicolumn{1}{c}{0.749} & 0.105 & \multicolumn{1}{c}{0.564} & \multicolumn{1}{c}{0.659} & 0.161 & \multicolumn{1}{c}{0.482} & \multicolumn{1}{c}{0.619} & 0.172 & \multicolumn{1}{c}{0.370} & \multicolumn{1}{c}{0.565} & 0.184 \\
    MB+M~\cite{zhang2015minimum} & 2015  & \multicolumn{1}{c}{0.470} & \multicolumn{1}{c}{0.597} & 0.177 & \multicolumn{1}{c}{0.554} & \multicolumn{1}{c}{0.618} & 0.146 & \multicolumn{1}{c}{0.692} & \multicolumn{1}{c}{0.726} & 0.129 & \multicolumn{1}{c}{0.487} & \multicolumn{1}{c}{0.609} & 0.206 & \multicolumn{1}{c}{0.562} & \multicolumn{1}{c}{0.661} & 0.158 & \multicolumn{1}{c}{0.342} & \multicolumn{1}{c}{0.538} & 0.228 \\
    RWRV~\cite{kim2015spatiotemporal} & 2015  & \multicolumn{1}{c}{0.345} & \multicolumn{1}{c}{0.556} & 0.199 & \multicolumn{1}{c}{0.438} & \multicolumn{1}{c}{0.583} & 0.162 & \multicolumn{1}{c}{0.440} & \multicolumn{1}{c}{0.595} & 0.188 & \multicolumn{1}{c}{0.336} & \multicolumn{1}{c}{0.521} & 0.242 & \multicolumn{1}{c}{0.422} & \multicolumn{1}{c}{0.552} & 0.211 & \multicolumn{1}{c}{0.283} & \multicolumn{1}{c}{0.504} & 0.245 \\
    SPVM~\cite{liu2014superpixel} & 2014  & \multicolumn{1}{c}{0.390} & \multicolumn{1}{c}{0.592} & 0.146 & \multicolumn{1}{c}{0.618} & \multicolumn{1}{c}{0.668} & 0.108 & \multicolumn{1}{c}{0.700} & \multicolumn{1}{c}{0.724} & 0.133 & \multicolumn{1}{c}{0.330} & \multicolumn{1}{c}{0.515} & 0.209 & \multicolumn{1}{c}{0.351} & \multicolumn{1}{c}{0.511} & 0.223 & \multicolumn{1}{c}{0.358} & \multicolumn{1}{c}{0.538} & 0.202 \\
    TIMP~\cite{zhou2014time} & 2014  & \multicolumn{1}{c}{0.448} & \multicolumn{1}{c}{0.593} & 0.172 & \multicolumn{1}{c}{0.573} & \multicolumn{1}{c}{0.644} & 0.116 & \multicolumn{1}{c}{0.479} & \multicolumn{1}{c}{0.612} & 0.170 & \multicolumn{1}{c}{0.456} & \multicolumn{1}{c}{0.576} & 0.192 & \multicolumn{1}{c}{0.401} & \multicolumn{1}{c}{0.575} & 0.215 & \multicolumn{1}{c}{0.395} & \multicolumn{1}{c}{0.563} & 0.195 \\
    SIVM~\cite{rahtu2010segmenting} & 2010  & \multicolumn{1}{c}{0.450} & \multicolumn{1}{c}{0.557} & 0.212 & \multicolumn{1}{c}{0.581} & \multicolumn{1}{c}{0.605} & 0.251 & \multicolumn{1}{c}{0.522} & \multicolumn{1}{c}{0.606} & 0.197 & \multicolumn{1}{c}{0.426} & \multicolumn{1}{c}{0.545} & 0.236 & \multicolumn{1}{c}{0.439} & \multicolumn{1}{c}{0.558} & 0.217 & \multicolumn{1}{c}{0.298} & \multicolumn{1}{c}{0.486} & 0.288 \\
    \bottomrule
    \end{tabular}%
   }
  \label{tab:Qualitative}%
\end{table*}%

\section{Experiments}
\subsection{Datasets and Evaluation Criteria}
\subsubsection{Evaluation Datasets.}
To evaluate the performance of our method,
we have conducted extensive quantitative evaluations over 6 widely used public benchmark datasets, including
DAVIS-T~\cite{perazzi2016benchmark}, SegTrack-V2~\cite{li2013video}, Visal~\cite{wang2015consistent},
FBMS-T~\cite{ochs2013segmentation}, VOS-T~\cite{li2017benchmark} and DAVSOD-T~\cite{fan2019shifting}.

\subsubsection{Evaluation Metrics.}
We use 3 widely adopted standard metrics in our quantitative evaluations:
F-measure~\cite{achanta2009frequency};
Structure Measure (S-measure)~\cite{fan2017structure};
Mean Absolute Error (MAE)~\cite{perazzi2012saliency}.

\begin{table*}[htbp]
  \centering
  \caption{Summarizing of the current SOTA video saliency detection training sets. DAVIS~\cite{perazzi2016benchmark}, SegTrack-V2~\cite{li2013video}, FBMS~\cite{ochs2013segmentation},  and DAVSOD~\cite{fan2019shifting}, MSRA10K~\cite{cheng2014global}, DUTOMRON~\cite{yang2013saliency}, PASCAL-S~\cite{li2014secrets}, HKU-IS~\cite{li2015visual}, DUTS~\cite{wang2017learning}, "---" indicates this dataset was not adopted in this methods.}
  \resizebox{\linewidth}{!}{
  \Huge
  \begin{tabular}{l|ccc}
    \toprule[1.5pt]
    Data  & OUR   & MGA19~\cite{li2019motion} & SSAV19~\cite{fan2019shifting} \\
    \midrule
    Videos & DAVSOD(5.5K)+DAVIS(2K)=7.5K & DAVIS(2K)+FBMS(0.5K)=2.5K & DAVSOD(5.5K)+DAVIS(2K)=7.5K \\
    Images & MSRA10K(4K)+HKU-IS(3K)+DUTOMRON(2.5K)=9.5K & DUTS(10.5K) & DUTOMRON(5K) \\
    Fixation & - & - & DAVSOD(5K) \\
    \midrule
    Total & Videos(7.5K)+Images(9.5K)=17K & Videos(2.5K)+Images(10.5K)=13K & Videos(7.5K)+Images(5K)+Fixation(5K)=17.5K
    \\
    \midrule
    \midrule
    Data  & PDBM18~\cite{song2018pyramid} & AGS19~\cite{wang2019learning} & DLVS18~\cite{wang2017video} \\
    \midrule
    Videos & DAVIS(2K) & - & SegTrack-V2(1K)+FBMS(0.5K)=1.5K \\
    Images & MSRA10K(10K)+DUTOMRON(5K)=15K & DUTOMRON(5K)+PASCAL-S:(1K)=6K & MSRA10K(10K)+DUTOMRON(5K)=15K \\
    Fixation & - & DAVIS(5.5K)+SegTrack-V2(1K)=6.5K & - \\
    \midrule
    Total & Videos(2K)+Images(15K)=17K & Images(6K)+Fixation(6.5K)=12.5K & Videos(1.5K)+Images(15K)=16.5K \\
    \bottomrule[1.5pt]
    \end{tabular}%
    }
  \label{tab:TrainingDatasets}%
\end{table*}%

\subsubsection{Training Set.}
Since the video saliency detection requires much more training data than the conventional image saliency detection, previous works~\cite{li2019motion, song2018pyramid, fan2019shifting} have followed the stage-wise training protocol as: pre-train the video saliency deep model using image data first and fine-tune it using video data latter.
As shown in Tab.~\ref{tab:TrainingDatasets}, we have listed the detailed training sets adopted by the current SOTA methods.
We have pre-trained our model using 9.5K image data, and these 9.5K images are selected from the DUTOMRON (2.5K)~\cite{yang2013saliency}, HKU-IS (3K)~\cite{li2016visual} and MSRA10K (4K)~\cite{cheng2014global}, in which we have removed those images without containing any mobilizable objects.
Then, we fine-tune our model using 7.5K video data, including the widely used DAVIS-TR (2K)~\cite{perazzi2016benchmark} and the recently proposed DAVSOD (5.5K)~\cite{fan2019shifting}.
It also should be noted that our training did not includes the fixation data of the DAVSOD dataset.

\subsubsection{Training Details.}
We firstly use the entire training set (all 17.5K data including both images \& videos) to pre-train our spatial branch (33,000 epoches).
All images/frames are resized to 256$\times$256, and we empirically set the batch size to 16.
Next, based on the above pre-trained models, we train the whole spatiotemporal model using the above training set (including both images \& videos).
Since our spatiotemporal network takes 3 frames as input each time, each static image was copied three times to meet the input size requirement.
And this training stage takes almost 8500 epoches, and we decrease the batch size from 16 to 4 here, because the spatiotemporal training takes more GPU memory.

Our network training uses stochastic gradient descent (SGD) with a momentum value of 0.9 and the weight decay is 5e-4,
and we set the initial learning rate 5e-3.
For relieving overfitting, we use random horizontal flip to augment the image training set.
Meanwhile, we re-sample the video data to have different frame rate using interval \{0,1,2,3,4,5,6\} to augment the video training set.
\begin{figure*}[htbp]
\centering
\includegraphics[width=1\linewidth]{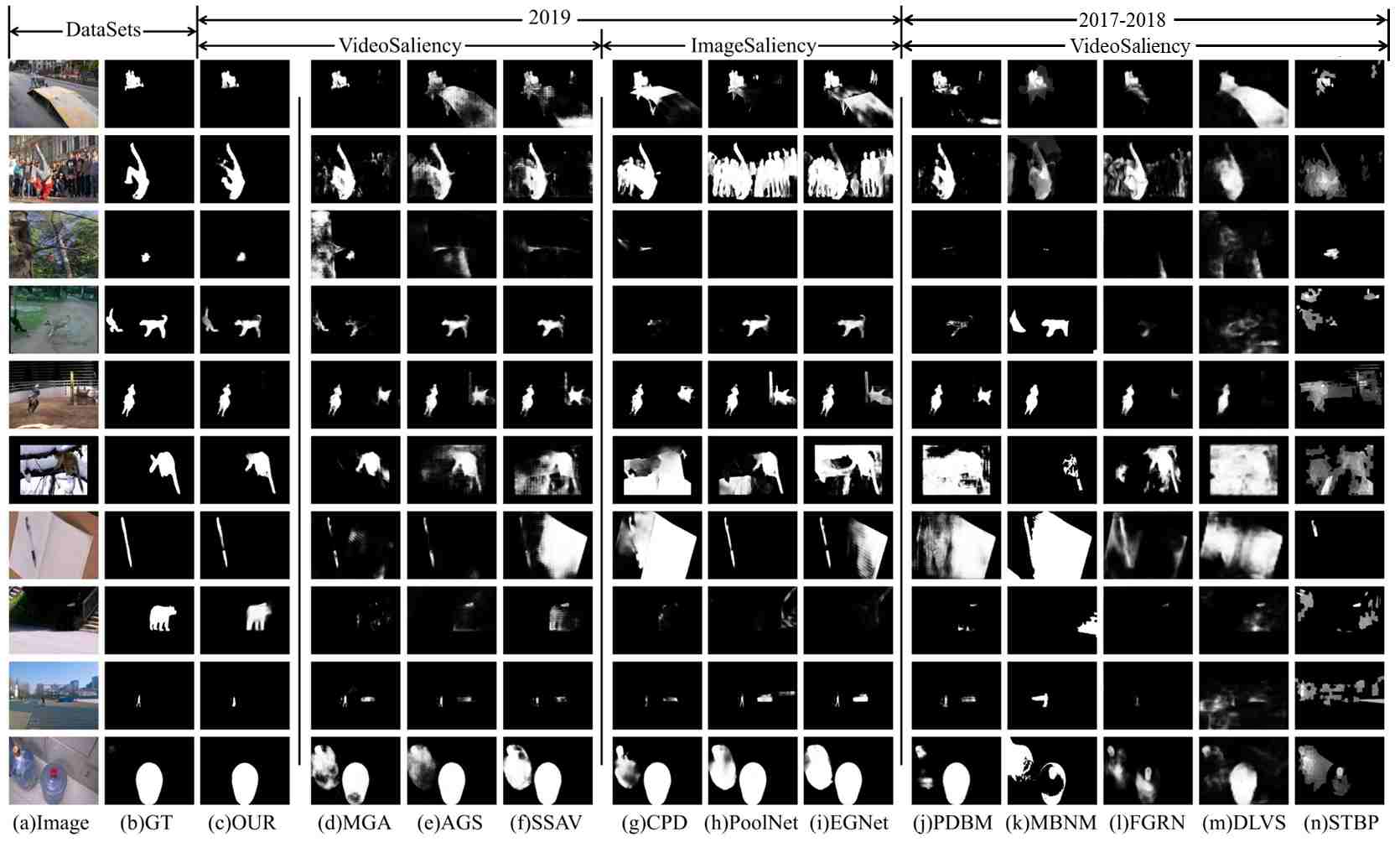}
  \vspace{-0.6cm}
\caption{Qualitative comparison with the SOTA video saliency detection and image saliency detection methods, where (a) Image donates the original video frame, (b) represents the human well-annotated VSOD ground truth, (c) is the VSOD results of our proposed method, (d) MGA~\cite{li2019motion}, (e) AGS~\cite{wang2019learning}, (f) SSAV~\cite{fan2019shifting}, (g) CPD~\cite{wu2019cascaded}, (h) PoolNet~\cite{liu2019simple}, (i) EGNet~\cite{zhao2019egnet}, (j) PDBM~\cite{song2018pyramid}, (k) MBNM~\cite{li2018unsupervised}, (l) FGRN~\cite{li2018flow}, (m) DLVS~\cite{wang2017video}, (n) STBP~\cite{xi2016salient}.}
\label{fig:Demo}
\end{figure*}

\subsection{Performance Comparisons}
We have compared our method with 23 SOTA saliency detection methods,
especially, 20 of which are video saliency detection methods,
MGA~\cite{li2019motion}, AGS~\cite{wang2019learning}, SSAV~\cite{fan2019shifting},
PDBM~\cite{song2018pyramid}, MBNM~\cite{li2018unsupervised},
FGRN~\cite{li2018flow}, DLVS~\cite{wang2017video}, SCNN~\cite{tang2018weakly},
SCOM~\cite{chen2018scom}, SFLR~\cite{chen2017video}, SGSP~\cite{liu2016saliency},
STBP~\cite{xi2016salient}, MSTM~\cite{tu2016real}, GFVM~\cite{wang2015consistent},
SAGM~\cite{wang2015saliency}, MB+M~\cite{zhang2015minimum}, RWRV~\cite{kim2015spatiotemporal},
SPVM~\cite{liu2014superpixel}, TIMP~\cite{zhou2014time} and SIVM~\cite{rahtu2010segmenting}.
We also compared our method with the most recent 3 SOTA image saliency detection methods,
CPD~\cite{wu2019cascaded}, PoolNet~\cite{liu2019simple}, EGNet~\cite{zhao2019egnet}.

\subsubsection{Quantitative Evaluation}
We employ three common metrics of F-max, S-measure, MAE for quantitative evaluation, and the Tab.~\ref{tab:Qualitative} shows the comparison details.
As shown in Tab.~\ref{tab:Qualitative}, our method basically maintains
the top three in all these tested datasets, especially,
our method shows the best performance on the SegTrack-V2 and VOS-T datasets.

It may be possible for our method to achieve more competitive results, e.g., we may achieve the best performances in FBMS and DAVSOD datasets if we include the FBMS-T and the Fixations into our training set as MGA~\cite{li2019motion} and SSAV~\cite{fan2019shifting} methods.
However, our key foci is to design a general video saliency framework with extremely fast speed (the highest FPS), it may not be very necessary to pursue the full-leading performance in all tested datasets.

Meanwhile, we have also provided a brief qualitative comparison in Fig.~\ref{fig:Demo}, showing the advantage of our method to handle video scenes with clutter backgrounds with moderate motions, e.g., the \emph{breakdance} sequence demonstrated in the 2nd row.

\subsubsection{Discussion}
In fact, the human fixation is extremely important to shift human attention between different objects.
Because the AGS~\cite{wang2019learning} has trained its model using the fixation data provided by the DAVSOD dataset, the AGS model has achieved the leading performance over the DAVSOD-T dataset.
On the other hand, the MGA~\cite{li2019motion} has achieved the best performance on the DAVIS-T dataset, because it has adopted the powerful off-the-shelf FlowNet2.0~\cite{ilg2017flownet}, which was pre-trained using massive additional training data, to sense temporal information, and the DAVIS-T dataset is dominated by motion information.
It should be noted that the leading performance of MGA over the FBMS dataset is mainly induced by its usage of the FBMS training set during the network training.
As for our spatiotemporal model, it has achieved the leading performance over the SegTrack-V2 and VOS-T datasets, in which these two datasets are respectively dominated by temporal and spatial information, showing the robustness of our method.
Also, we have provided the qualitative comparisons in Fig.~\ref{fig:Demo}.

\subsubsection{Efficiency Comparison.}
We also report the runtime comparisons and the net size comparisons in Tab.~\ref{tab:Runtime} and Tab.~\ref{tab:Weight}.
Our method is evaluated on a machine with a GTX1080Ti GPU.
As shown in Tab.~\ref{tab:Runtime}, thought our method has evaluated on a GTX1080Ti GPU, it has achieved the highest FPS compared to all the other compared models even in the case that several other models are evaluated on the more powerful GPUs.
\begin{table}[htbp]
  \centering
  \caption{Runtime comparisons between our method and multiple most representative SOTA video saliency detection methods.}
  \setlength{\tabcolsep}{7mm}{
  \resizebox{\linewidth}{!}{
    \large
    \begin{tabular}{l|ccc}
    \toprule[1pt]
    Methods & OUR   & MGA19~\cite{li2019motion}  & AGS19~\cite{wang2019learning}  \\
    \midrule
    FPS   & \textcolor[rgb]{1.00,0.00,0.00}{50.0}  & \textcolor[rgb]{0.00,0.00,1.00}{14.0}  & 10.0  \\
    Platform  & GTX1080Ti  & GTX1080Ti  & GTX1080Ti  \\
    \midrule
    \midrule
    Methods & COS19~\cite{lu2019see} &  SSAV19~\cite{fan2019shifting}  & PDBM18~\cite{song2018pyramid} \\
    \midrule
    FPS   & 0.4   & \textcolor[rgb]{0.00,1.00,0.00}{20.0}  &  \textcolor[rgb]{0.00,1.00,0.00}{20.0} \\
    Platform  & GTX1080Ti &  GTXTITANX &  GTXTITANX \\
    \bottomrule[1pt]
    \end{tabular}%
    }
    }
  \label{tab:Runtime}%
    \vspace{-0.2cm}
\end{table}%

Tab.~\ref{tab:Weight} demonstrates the network parameter amount comparisons between our model and multiple most representative SOTA models, in which our method is with the lightest network architecture.

\begin{table}[htbp]
  \centering
  \caption{Net size comparisons between our method with the multiple most representative SOTA video saliency detection methods.}
  \setlength{\tabcolsep}{7mm}{
  \resizebox{\linewidth}{!}{
  \large
    \begin{tabular}{l|ccc}
    \toprule[1pt]
    Methods & OUR   & MGA19~\cite{li2019motion}  & AGS19~\cite{wang2019learning}  \\
    \midrule
    Weight(M) & \textcolor[rgb]{1.00,0.00,0.00}{191.0}  & 349.8  & 262.0  \\
    Toolbox & Pytorch & Pytorch & Caffe \\
    \midrule
    \midrule
    Methods & COS19~\cite{lu2019see} &  SSAV19~\cite{fan2019shifting}  & PDBM18~\cite{song2018pyramid} \\
    \midrule
    Weight(M) & 310.5  & \textcolor[rgb]{0.00,0.00,1.00}{236.2}  & \textcolor[rgb]{0.00,1.00,0.00}{236.0}  \\
    Toolbox & Pytorch & Caffe & Caffe \\
    \bottomrule[1pt]
    \end{tabular}%
    }
    }
  \label{tab:Weight}%
\end{table}%

\begin{table}[!t]
  \centering
  \caption{Component evaluation of our method on 6 datasets.``+3D'' represents the performance after using the fast 3D convolution model (Sec.~\ref{sec:S}), ``+S'' represents the performance after using the fast temporal shuffle model (Sec.~\ref{sec:3D}), ``+MA'' represents the performance after using the multi-scale attention (Sec.~\ref{sec:Flow}).}
  \resizebox{\linewidth}{!}{
    \begin{tabular}{l|ccc|ccc}
    \toprule[1pt]
    DataSets & \multicolumn{3}{c|}{DAVIS-T~\cite{perazzi2016benchmark}} & \multicolumn{3}{c}{SegTrack-V2~\cite{li2013video}} \\
    \midrule
    Metric & F-max & S-measure & MAE   & F-max & S-measure & MAE \\
    \midrule
    +3D+S+MA & 0.865  & 0.892  & 0.023  & 0.860  & 0.891  & 0.017  \\
    +3D+S & 0.858  & 0.890  & 0.024  & 0.855  & 0.891  & 0.017  \\
    +3D   & 0.855  & 0.895  & 0.027  & 0.841  & 0.884  & 0.017  \\
    Baseline & 0.837  & 0.878  & 0.032  & 0.822  & 0.876  & 0.022  \\
    \midrule
    \midrule
    DataSets & \multicolumn{3}{c|}{Visal~\cite{wang2015consistent}} & \multicolumn{3}{c}{FBMS-T~\cite{ochs2013segmentation}} \\
    \midrule
    Metric & F-max & S-measure & MAE   & F-max & S-measure & MAE \\
    \midrule
    +3D+S+MA & 0.952  & 0.952  & 0.013  & 0.856  & 0.872  & 0.038  \\
    +3D+S & 0.951  & 0.949  & 0.017  & 0.852  & 0.871  & 0.039  \\
    +3D   & 0.942  & 0.943  & 0.016  & 0.853  & 0.870  & 0.038  \\
    Baseline & 0.912  & 0.924  & 0.025  & 0.839  & 0.867  & 0.047  \\
    \midrule
    \midrule
    DataSets & \multicolumn{3}{c|}{VOS-T~\cite{li2017benchmark}} & \multicolumn{3}{c}{DAVSOD~\cite{fan2019shifting}} \\
    \midrule
    Metric & F-max & S-measure & MAE   & F-max & S-measure & MAE \\
    \midrule
    +3D+S+MA & 0.791  & 0.850  & 0.058  & 0.651  & 0.746  & 0.086  \\
    +3D+S & 0.782  & 0.847  & 0.062  & 0.650  & 0.747  & 0.085  \\
    +3D   & 0.791  & 0.851  & 0.060  & 0.650  & 0.748  & 0.086  \\
    Baseline & 0.771  & 0.839  & 0.062  & 0.629  & 0.725  & 0.099  \\
    \bottomrule[1pt]
    \end{tabular}%
    }
  \label{tab:Component}%
\end{table}%

\subsection{Component Evaluation}
In Tab.~\ref{tab:Component}, we have conducted multiple component evaluations to verify the effectiveness of each component, in which the baseline represents the original spatial branch using 2D convolution only.

Regarding our temporal shuffle, it will introduce noisy information indeed, and its effect reported in Tab.~\ref{tab:Component} is not significant even in the case where the temporal sensing ability has been enhanced via temporal shuffle.
We choose to use the attention model to compensate the lost location information during temporal shuffle, and the performance improvement after using the attention model (i.e., 3D+S+MA) is significant.

\begin{figure}[t]
\centering
\vspace{0.3cm}
\includegraphics[width=1\linewidth]{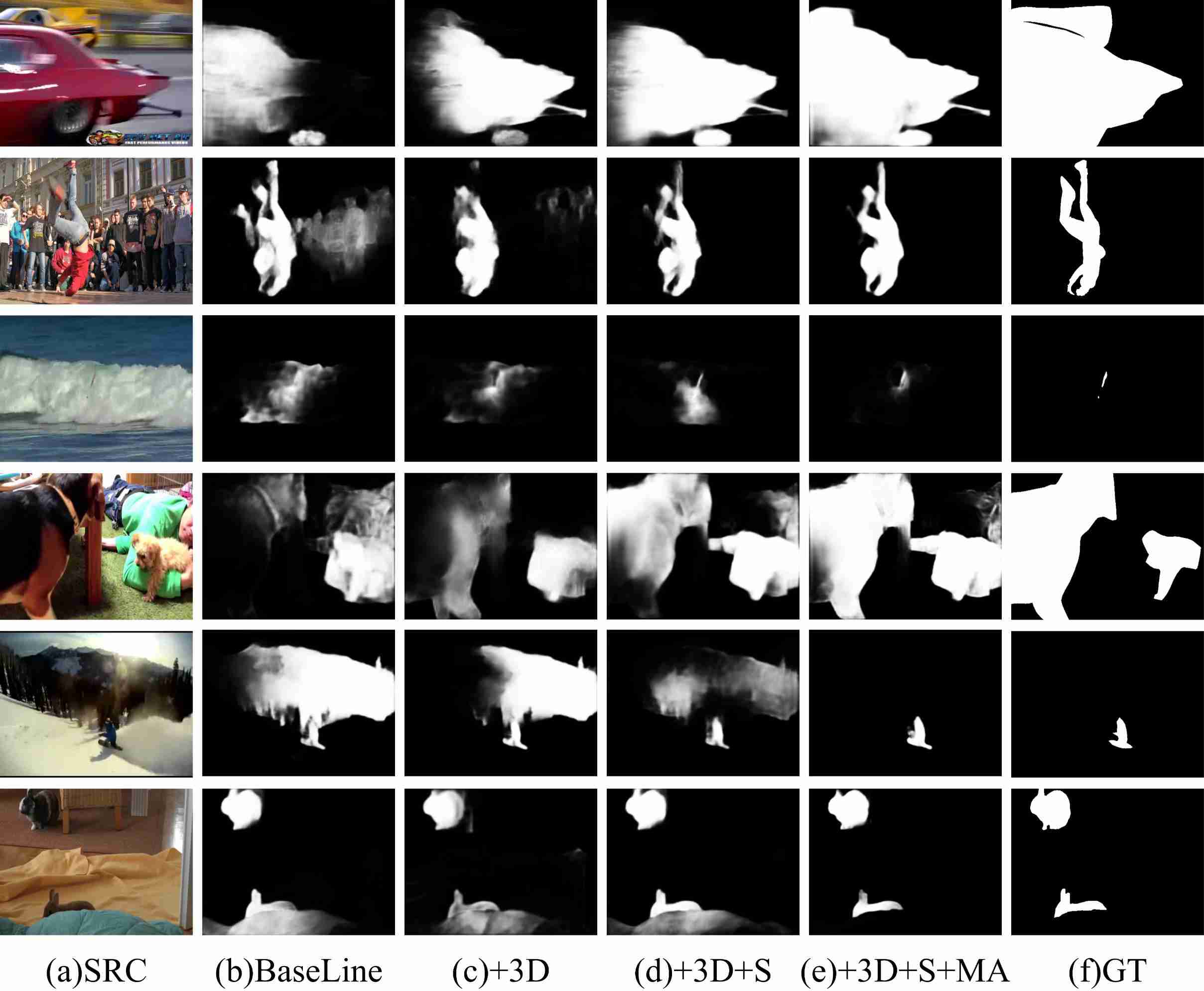}
\caption{Qualitative demonstrations of each major component in our method; ``+3D'' represents the saliency maps after using the fast 3D convolution model (Sec.~\ref{sec:S}); ``+S'' represents the saliency maps after using the fast temporal shuffle model (Sec.~\ref{sec:3D}); ``+MA'' represents the saliency maps after using the multi-scale attention (Sec.~\ref{sec:Flow}).}
  \vspace{-0.4cm}
\label{fig:ComponentDemo}
\end{figure}

The qualitative comparisons between different components can be found in Fig.~\ref{fig:ComponentDemo}.

\subsubsection{\textbf{Effectiveness of 3D Convolution (Sec.~\ref{sec:3D})}}
As shown in Tab.~\ref{tab:Component}, the solely 2D convolution based baseline shows the worst performance in all these tested datasets.
Benefited by the 3D convolution provided temporal information, the overall performance of the baseline network achieves averagely 2\% improvements as expected.

\begin{table}[htbp]
  \centering
  \caption{The ablation study toward the number of sequential 3D convolutions in temporal model (i.e., 3D\_Ri, in which the $i\in\{1,3,5\}$ denotes the 3D convolution number, ) increasing improvement for performance, the best scores are labeled in \textcolor[rgb]{1.00,0.00,0.00}{red}.}
  \resizebox{\linewidth}{!}{
    \begin{tabular}{l|ccc|ccc}
    \toprule[1pt]
    DataSets & \multicolumn{3}{c|}{DAVIS-T~\cite{perazzi2016benchmark}} & \multicolumn{3}{c}{SegTrack-V2~\cite{li2013video}} \\
    \midrule
    Metric & F-max & S-measure & MAE   & F-max & S-measure & MAE \\
    \midrule
    3D\_R5 & 0.841  & 0.882  & 0.025  & 0.832  & \textcolor[rgb]{ 1,  0,  0}{0.890} & \textcolor[rgb]{ 1,  0,  0}{0.015} \\
    3D\_R3 & \textcolor[rgb]{ 1,  0,  0}{0.855} & 0.895  & \textcolor[rgb]{ 1,  0,  0}{0.027} & 0.841  & 0.884  & 0.017  \\
    3D\_R1 & 0.839  & \textcolor[rgb]{ 1,  0,  0}{0.896} & 0.028  & \textcolor[rgb]{ 1,  0,  0}{0.847} & 0.873  & 0.021  \\
    Baseline & 0.837  & 0.878  & 0.032  & 0.822  & 0.876  & 0.022  \\
    \midrule
    \midrule
    DataSets & \multicolumn{3}{c|}{Visal~\cite{wang2015consistent}} & \multicolumn{3}{c}{FBMS-T~\cite{ochs2013segmentation}} \\
    \midrule
    Metric & F-max & \multicolumn{1}{r}{S-measure} & MAE   & F-max & S-measure & MAE \\
    \midrule
    3D\_R5 & 0.937  & 0.941  & \textcolor[rgb]{ 1,  0,  0}{0.013} & 0.837  & 0.866  & 0.041  \\
    3D\_R3 & \textcolor[rgb]{ 1,  0,  0}{0.942} & 0.943  & 0.016  & \textcolor[rgb]{ 1,  0,  0}{0.853} & \textcolor[rgb]{ 1,  0,  0}{0.870} & \textcolor[rgb]{ 1,  0,  0}{0.038} \\
    3D\_R1 & 0.921  & \textcolor[rgb]{ 1,  0,  0}{0.948} & 0.018  & 0.851  & 0.869  & 0.043  \\
    Baseline & 0.912  & 0.924  & 0.025  & 0.839  & 0.867  & 0.047  \\
    \midrule
    \midrule
    DataSets & \multicolumn{3}{c|}{VOS-T~\cite{li2017benchmark}} & \multicolumn{3}{c}{DAVSOD~\cite{fan2019shifting}} \\
    \midrule
    Metric & F-max & S-measure & MAE   & F-max & S-measure & MAE \\
    \midrule
    3D\_R5 & 0.780  & 0.842  & \textcolor[rgb]{ 1,  0,  0}{0.055} & 0.647  & 0.742  & \textcolor[rgb]{ 1,  0,  0}{0.083} \\
    3D\_R3 & \textcolor[rgb]{ 1,  0,  0}{0.791} & 0.851  & 0.060  & \textcolor[rgb]{ 1,  0,  0}{0.650} & \textcolor[rgb]{ 1,  0,  0}{0.748} & 0.086  \\
    3D\_R1 & 0.789  & \textcolor[rgb]{ 1,  0,  0}{0.853} & 0.061  & 0.648  & 0.744  & 0.087  \\
    Baseline & 0.771  & 0.839  & 0.062  & 0.629  & 0.725  & 0.099  \\
    \bottomrule[1pt]
    \end{tabular}%
   }
  \label{tab:STINFO}
\end{table}%

Meanwhile, we have conducted an ablation study toward the NUMBER of the adopted sequential 3D convolutions.
As shown in Tab.~\ref{tab:STINFO}, the overall performance can get improvement gradually when we increase the used sequential 3D convolutions, e.g., \rm{3D\_R(1$\rightarrow$3)}.
However, we have noticed a slight performance degeneration when we use 5 sequential 3D convolutions in our temporal model (i.e., $\rm{3D\_R5}$), which may be induced by the miss-aligned spatial information in these sequential 3D convolutions, leading to an extremely large problem domain for its spatiotemporal deep feature computation.
Therefore, we decide to use 3 sequential 3D convolutions in our method.

Also, we have compared the performance difference by using different padding schemes in our temporal model, in which we have listed the detailed performance via our cyclic padding and the conventional zero padding.
As shown in Tab.~\ref{tab:Padding}, we have noticed that our ``Cyclic Padding'' scheme can effectively boost the performance of our method mainly over the SegTrack-V2 dataset, while the overall performance improvement over the rest datasets are marginal.
This is mainly induced by the fact that, among all these tested datasets, only the SegTrack-V2 dataset is fully dominated by fast object movements, and the ``Cyclic Padding'' scheme can effectively robust the temporal model for the fast movement detection (e.g., the $birdfall$ sequence in SegTrack-V2 dataset), and thus we can achieve a significant performance improvement over the SegTrack-V2 dataset.

\begin{table}[htbp]
  \centering
    \vspace{-0.2cm}
  \caption{The padding scheme comparison in temporal model; "CyclicPadding" represents our proposed cyclic padding scheme and "ZeroPadding" represents the original zero padding.}
  \vspace{-0.2cm}
  \resizebox{\linewidth}{!}{
  \begin{tabular}{l|ccc|ccc}
    \toprule[1pt]
    DataSets & \multicolumn{3}{c|}{DAVIS-T~\cite{perazzi2016benchmark}} & \multicolumn{3}{c}{SegTrack-V2~\cite{li2013video}} \\
    \midrule
    Metric & F-max & S-measure & MAE   & F-max & S-measure & MAE \\
    \midrule
    CyclicPadding & 0.865  & 0.892  & 0.023  & 0.860  & 0.891  & 0.017  \\
    ZeroPadding & 0.857  & 0.889  & 0.024  & 0.848  & 0.890  & 0.018  \\
    \midrule
    \midrule
    DataSets & \multicolumn{3}{c|}{Visal~\cite{wang2015consistent}} & \multicolumn{3}{c}{FBMS-T~\cite{ochs2013segmentation}} \\
    \midrule
    Metric & F-max & S-measure & MAE   & F-max & S-measure & MAE \\
    \midrule
    CyclicPadding & 0.952  & 0.952  & 0.013  & 0.856  & 0.872  & 0.038  \\
    ZeroPadding & 0.938  & 0.944  & 0.016  & 0.852  & 0.871  & 0.039  \\
    \midrule
    \midrule
    DataSets & \multicolumn{3}{c|}{VOS-T~\cite{li2017benchmark}} & \multicolumn{3}{c}{DAVSOD~\cite{fan2019shifting}} \\
    \midrule
    Metric & F-max & S-measure & MAE   & F-max & S-measure & MAE \\
    \midrule
    CyclicPadding & 0.791  & 0.850  & 0.058  & 0.651  & 0.746  & 0.086  \\
    ZeroPadding & 0.790  & 0.848  & 0.058  & 0.649  & 0.745  & 0.088  \\
    \bottomrule[1pt]
    \end{tabular}%
    }
  \label{tab:Padding}%
\end{table}%

\subsubsection{\textbf{Effectiveness of Temporal Shuffle (Sec.~\ref{sec:S})}}
As shown in Tab.~\ref{tab:Component}, the overall performance can get further improved by integrating the temporal shuffle scheme into our temporal model.
The major highlight of the temporal shuffle is that it integrates the temporally neighbored spatial information into the current spatial domain, which biases the adopted 3D kernel to the temporal scale.
For an example, the temporal shuffle operation is able to enhance the temporal information sensing ability of the temporal model, and thus the overall performance can get a large performance improvements over those temporal information dominated datasets, e.g., DAVIS-T and SegTrack-V2 datasets.

\subsubsection{\textbf{Effectiveness of Multi-scale Attention (Eq.~\ref{eq:STFeature})}}
As shown in Tab.~\ref{tab:Component}, the overall performance can get further improved by integrating the multi-scale attention into each decoder layer.
Thought the attention has been adopted by the most recent work~\cite{fan2019shifting} in the single-scale manner, here, our multi-scale spatiotemporal deep feature computation has integrated such attention in the multi-scale way, which further improves the overall detection performance.

\begin{table}[htbp]
  \centering
    \vspace{-0.2cm}
  \caption{The performance comparisons between different decoder layers; ${\rm DeCoder}\_i, i\in\{5,4,3,2,1\}$ represents the output of the $i$-th decoder layer respectively, see more details in Fig.~\ref{fig:Net}.}
  \vspace{-0.2cm}
  \resizebox{\linewidth}{!}{
    \begin{tabular}{l|ccc|ccc}
    \toprule[1pt]
    DataSets & \multicolumn{3}{c|}{DAVIS-T~\cite{perazzi2016benchmark}} & \multicolumn{3}{c}{SegTrack-V2~\cite{li2013video}} \\
    \midrule
    Metric & F-max & S-measure & MAE   & F-max & S-measure & MAE \\
    \midrule
    DeCoder\_1 & 0.865  & 0.892  & 0.023  & 0.860  & 0.891  & 0.017  \\
    DeCoder\_2 & 0.863  & 0.891  & 0.022  & 0.857  & 0.891  & 0.016  \\
    DeCoder\_3 & 0.861  & 0.890  & 0.024  & 0.853  & 0.885  & 0.017  \\
    DeCoder\_4 & 0.839  & 0.874  & 0.026  & 0.827  & 0.863  & 0.020  \\
    DeCoder\_5 & 0.806  & 0.855  & 0.032  & 0.792  & 0.839  & 0.024  \\
    \midrule
    \midrule
    DataSets & \multicolumn{3}{c|}{Visal~\cite{wang2015consistent}} & \multicolumn{3}{c}{FBMS-T~\cite{ochs2013segmentation}} \\
    \midrule
    Metric & F-max & S-measure & MAE   & F-max & S-measure & MAE \\
    \midrule
    DeCoder\_1 & 0.952  & 0.952  & 0.013  & 0.856  & 0.872  & 0.038  \\
    DeCoder\_2 & 0.948  & 0.949  & 0.013  & 0.853  & 0.782  & 0.038  \\
    DeCoder\_3 & 0.942  & 0.945  & 0.015  & 0.848  & 0.868  & 0.040  \\
    DeCoder\_4 & 0.923  & 0.932  & 0.018  & 0.832  & 0.854  & 0.044  \\
    DeCoder\_5 & 0.899  & 0.915  & 0.024  & 0.807  & 0.833  & 0.050  \\
    \midrule
    \midrule
    DataSets & \multicolumn{3}{c|}{VOS-T~\cite{li2017benchmark}} & \multicolumn{3}{c}{DAVSOD~\cite{fan2019shifting}} \\
    \midrule
    Metric & F-max & S-measure & MAE   & F-max & S-measure & MAE \\
    \midrule
    DeCoder\_1 & 0.791  & 0.850  & 0.058  & 0.651  & 0.746  & 0.086  \\
    DeCoder\_2 & 0.790  & 0.849  & 0.058  & 0.651  & 0.746  & 0.086  \\
    DeCoder\_3 & 0.787  & 0.847  & 0.058  & 0.648  & 0.744  & 0.087  \\
    DeCoder\_4 & 0.769  & 0.836  & 0.059  & 0.639  & 0.736  & 0.087  \\
    DeCoder\_5 & 0.745  & 0.816  & 0.064  & 0.619  & 0.725  & 0.090  \\
    \bottomrule[1pt]
    \end{tabular}%
   }
   \vspace{0.2cm}
  \label{tab:MultiScale}%
\end{table}%

\subsubsection{\textbf{Effectiveness of Multi-scale Spatiotemporal Recurrent (Eq.~\ref{eq:STFeatureFull})}}
In order to verify the effectiveness of our spatiotemporal recurrent, we have respectively tested the performance of each decoder layer, i.e., $DeCoder_i, i\in\{5,4,3,2,1\}$ in Fig.~\ref{fig:Net}.
As shown in Tab.~\ref{tab:MultiScale}, the last recurrent layer $DeCoder_1$ has achieved the best result.

\subsubsection{\textbf{The Reasons Why We Choose A Small Feature Size For Our Spatial-branch}}
In SSAV~\cite{fan2019shifting} and PDBM~\cite{song2018pyramid}, the interactions between their
spatial and temporal branches are limited, which solely feed the last
output of spatial branch into temporal branch, and such
``single-scale interaction'' has one critical weakness, i.e., their
final detections are frequently associated with blurred object boundaries,
which is mainly induced by their temporal branches (i.e., ConvLSTM).
Therefore, these two works resort to the attention model to compensate
the lost object boundaries, and their dilated convolutions must choose a
relatively large size (60*60) in general to ensure the effectiveness of
their attention models.

In sharp contrast to SSAV and PDBM, our method introduces the ``multi-scale''
spatial information into temporal branch by using side-outputs of different
spatial layers (Fig.~\ref{fig:Net}), which ensures the detection results with sharp
object boundaries.
The attention module adopted in our method aims to provide the
``location information'' for temporal branch, thus it is totally acceptable to design our spatial-branch with a small feature size.

%
\subsection{Limitations}
\begin{figure}[t]
\centering
\includegraphics[width=1\linewidth]{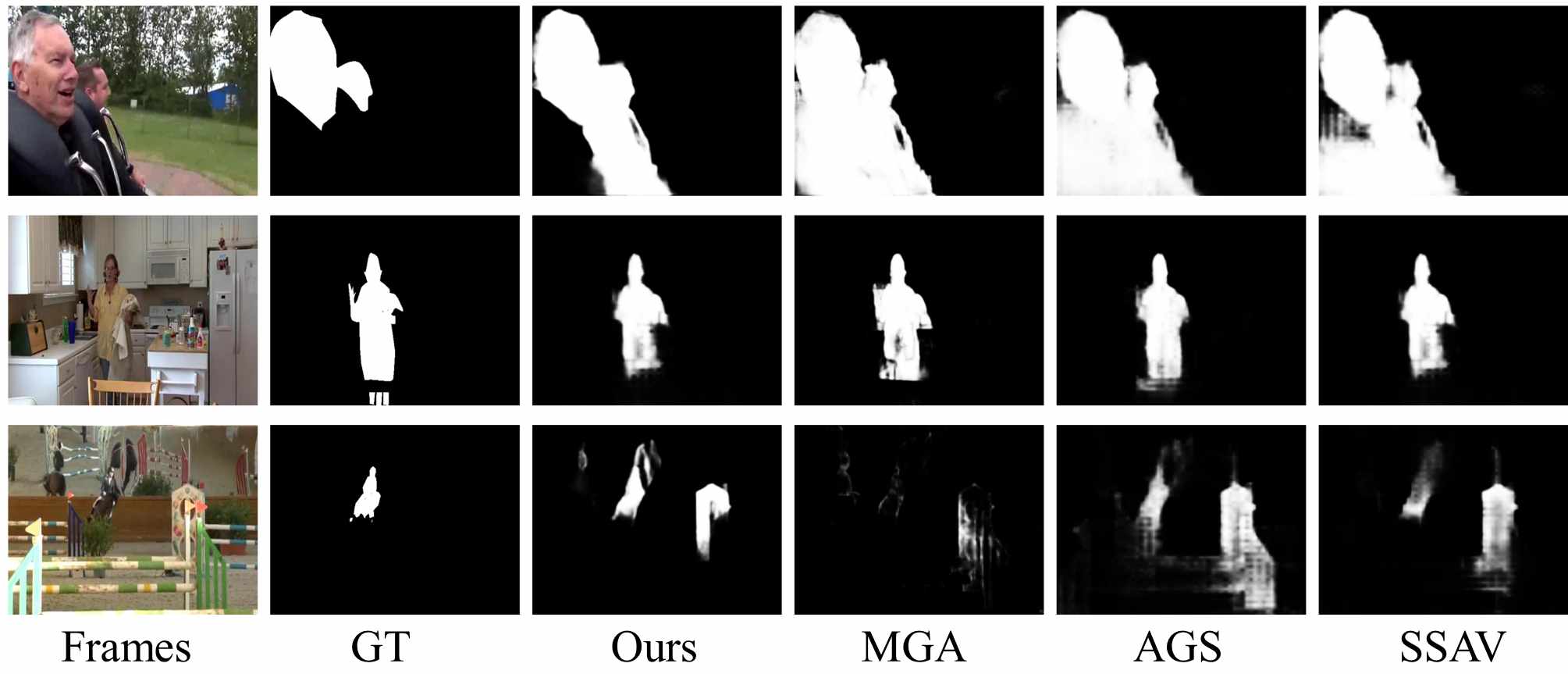}
\caption{The failure cases of our proposed method. Meanwhile, we have also provided other three most representative SOTA methods as references.}
\label{fig:Limitations}
\vspace{-0.2cm}
\end{figure}
Because our method only takes 3 consecutive video frames as input each time, its sensing scope for temporal information is quite limited, leading those regions which undergo long-period static status being undetected, see pictorial demonstrations in Fig.~\ref{fig:Limitations}.
Also, this limitation is also quite common in the SOTA methods, and we believe it will may be alleviated by introducing the long-term spatiotemporal information, which deserves our future investigation.

\section{Conclusion}
In this paper, we propose an extremely fast end-to-end video saliency detection method.
The major highlight of our method can be summarized into three solid aspects: \underline{1)} We have devised a lightweight temporal model, which can be inserted into each decoder layer to obtain the multi-scale spatiotemporal deep features;
\underline{2)} We have provided a feasible way to apply a sequential of 3D convolutions to sense the temporal information;
\underline{3)} We have introduced a fast temporal shuffle scheme to enhance the temporal sensing ability of the 3D convolution;
Also, we have conducted extensive quantitative evaluations to verbify the effectiveness of each component in our method.
And the quantitative comparisons have indicated that our method outperforms all the current SOTA methods in both detection performance and speed.


\vspace{-0.1cm}
\bibliographystyle{IEEEtran}
\bibliography{paper}

\end{document}